\documentclass{article}
\usepackage[hidelinks]{hyperref}
\usepackage{ijcai19}
\usepackage{amssymb}

\usepackage{times}
\usepackage{soul}
\usepackage{url}
\usepackage[utf8]{inputenc}
\usepackage[small]{caption}
\usepackage{graphicx}
\usepackage{amsmath}
\usepackage{booktabs}
\title{Conflict as an Inverse of Attention in Sequence Relationship \footnote{This work was presented at 2nd Workshop on Humanizing AI (HAI) at IJCAI'19 in Macao, China.}}

\author{
Rajarshee Mitra$^1$\footnote{Contact Author}
\affiliations
$^1$Microsoft\\
\emails
\
ramitra@microsoft.com
}

\begin{document}

\maketitle

\begin{abstract}
Attention is a very efficient way to model the relationship between two sequences by comparing how similar two intermediate representations are. Initially demonstrated in NMT, it is a standard in all NLU tasks today when efficient interaction between sequences is considered. However, we show that attention, by virtue of its composition, works best only when it is given that there is a match somewhere between two sequences. It does not very well adapt to cases when there is no similarity between two sequences or if the relationship is contrastive. We propose an Conflict model which is very similar to how attention works but which emphasizes mostly on how well two sequences repel each other and finally empirically show how this method in conjunction with attention can boost the overall performance.
\end{abstract}

\maketitle

\section{Introduction}
Modelling the relationship between sequences is extremely significant in most retrieval or classification problems involving two sequences. Traditionally, in Siamese networks, Hadamard product or concatenation have been used to fuse two vector representations of two input sequences to form a final representation for tasks like semantic similarity, passage retrieval. This representation, subsequently, has been used to compute similarity scores which has been used in a variety of training objectives like margin loss for ranking or cross-entropy error in classification.

We have also witnessed word or phrase level similarity to create alignment matrices between two sequences \cite{DBLP:conf/wmt/NiehuesV08}, \cite{DBLP:conf/semeval/SultanBS15}. These alignment matrices has proved to be very useful to model the relationship between two word representations as well fuse the relevant information of one sequence into another. Empirical evidences have shown this alignment procedures have significantly performed better then simple concatenation or element-wise multiplication, especially for long sentences or paragraphs. 

Attention works on creating neural alignment matrix using learnt weights without pre-computing alignment matrix and using them as features.
The main objective of any attentive or alignment process is to look for matching words or phrases between two sequences and assign a high weight to the most similar pairs and vice-versa. The notion of matching or similarity maybe not semantic similarity but based on whatever task we have at hand. For example, for a task that requires capturing semantic similarity between two sequences like "how rich is tom cruise" and "how much wealth does tom cruise have", an attentive model shall discover the high similarity between "rich" and "wealthy" and assign a high weight value to the pair. Likewise, for a different task like question answering, a word "long" in a question like "how long does it take to recover from a mild fever" might be aligned with the phrase "a week" from the candidate answer "it takes almost a week to recover fully from a fever".
Thus, attention significantly aids in better understanding the relevance of a similar user query in a similar measurement task or a candidate answer in a question answering task. The final prediction score is dependent on how well the relationship between two sequences are modeled and established.

The general process of matching one sequence with another through attention includes computing the alignment matrix containing weight value between every pair of word representations belonging to both of the sequences. Subsequently, softmax function is applied on all the elements of one of the two dimensions of the matrix to represent the matching probabilities of all the word of a sequence with respect to one particular word in the other sequence.  

Since attention always looks for matching word representations, it operates under the assumption that there is always a match to be found inside the sequences. We provide a theoretical limitation to it and propose another technique called \textit{conflict} that looks for contrasting relationship between words in two sequences. We empirically verify that our proposed conflict mechanism combined with attention can outperform the performance of attention working solely.

\section{Related Work} 
Bahdanau et al. \cite{DBLP:journals/corr/BahdanauCB14} introduced attention first in neural machine translation. It used a feed-forward network over addition of encoder and decoder states to compute alignment score. Our work is very similar to this except we use element wise difference instead of addition to build our conflict function. \cite{DBLP:conf/nips/VaswaniSPUJGKP17} came up with a scaled dot-product attention in their Transformer model which is fast and memory-efficient. Due to the scaling factor, it didn't have the issue of gradients zeroing out. On the other hand, \cite{DBLP:conf/emnlp/LuongPM15} has experimented with global and local attention based on the how many hidden states the attention function takes into account. Their experiments have revolved around three attention functions - dot, concat and general. Their findings include that dot product works best for global attention. Our work also belongs to the global attention family as we consider all the hidden states of the sequence. 

Attention has been widely used in pair-classification problems like natural language inference. Wang et al. \cite{DBLP:conf/ijcai/WangHF17} introduced BIMPM which matched one sequence with another in four different fashion but one single matching function which they used as cosine. Liu et al. \cite{DBLP:journals/corr/abs-1804-07888} proposed SAN for language inference which also used dot-product attention between the sequences. 

Summarizing, attention has helped in achieving state-of-the-art results in NLI and QA. Prior work in attention has been mostly in similarity based approaches while our work focuses on non-matching sequences.

\section{How attention works}

Let us consider that we have two sequences $u$ and $v$ each with \textbf{M} and \textbf{N} words respectively. The objective of attention is two-fold: compute alignment scores (or weight) between every word representation pairs from $u$ and $v$ and fuse the matching information of $u$ with $v$ thus computing a new representation of $v$ conditioned on $u$.

The word representations that attention operates on can be either embeddings like GloVe or hidden states from any recurrent neural network. We denote these representations as u = $\{w_t^u\}_{t=1}^M$ and v = $\{w_t^v\}_{t=1}^N$. We provide a mathematical working of how a general attention mechanism works between two sequences, followed by a explanation in words:
			\begin{align}
            u^{linear}&=tanh(W^{u}u); v^{linear}=tanh(W^{v}v) \\
			a_{ij} &= u_{i}^{linear}.{v_j^{linear}}^T \\
            w_i &= softmax(a_i) \\
            v^{weighted}&= w_i v \\
            u_{i}^{new}&=[u_i; v^{weighted}] \label{attention} \\
			\end{align}
\textbf{Explanation}: Both are sequences are non-linearly projected into two different spaces (eqn.1) and each word representation in $u$ is matched with that in $v$ by computing a dot-product (eqn.2). $a$ is a M X N matrix that stores the alignment scores between word $u_i$ and $v_j$ (eqn.2). Since, the scores are not normalized, a softmax function is applied on each row to convert them to probabilities (eqn. 3). Thus, each row contains relative importance of words in $v$ to a particular word $u_i$. Weighted sum of $v$ is taken (eqn. 4) and fused with the word representation $u_i$ using concatenation (eqn.5).
\begin{figure}[h]
  \centering
  \includegraphics[width=\linewidth,height=3in]{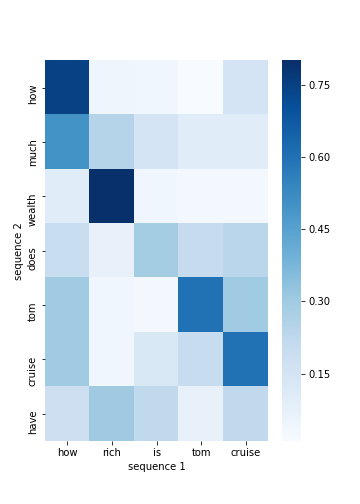}
  \caption{Attention Heatmaps}
\end{figure}
\begin{figure}[h]
  \centering
  \includegraphics[width=\linewidth,height=2in]{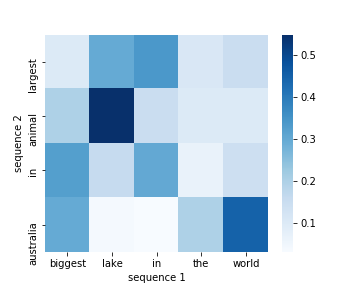}
  \caption{Conflict Heatmaps}
\end{figure}

\section{Limits of using only Attention}
Attention operates by using dot product or sometimes addition followed by linear projection to a scalar which models the similarity between two vectors. Subsequently, softmax is applied which gives high probabilities to most matching word representations. This assumes that there is some highly matched word pairs already existing and high scores will be assigned to them. Given a vector $x$=($x_1$,...,$x_n$) on which softmax function is applied, each $x_i$ $\in$ (0, 1). It is observable that the average value of $x_i$ is always $\frac{1}{n}$. In other words, it is impossible to produce a vector having all $x_i$ < $\frac{1}{n}$ when two sequences have no matching at all.

In cases, where one or more word pairs from two different sequences are highly dissimilar, it is impossible to assign a very low probability to it without increasing the probability of some other pair somewhere else since $\sum\nolimits_i x$ = 1.

For example, when we consider two sequences "height of tom cruise" and "age of sun", while computing the attention weights between the word "height" and all the words in the second sequence it can be observed that their no matching word in the latter. In this case, a standard dot-product based attention with softmax won't be able to produce weights which is below 0.33 (=1/3) for all the words in the second sequence with respect to the word "height" in the first sequence.

\section{Conflict model}
We propose a different mechanism that does the opposite of what attention does that is computing how much two sequences repel each other. This works very similar to how attention works but inversely. 

We demonstrate a general model but we also realize that there can be other variants of it which may be worked out to perform better. Our approach consists of using element wise difference between two vectors followed by a linear transformation to produce a scalar weight. The remaining of the process acts similar to how attention works. Mathematically, we can express it as:

			\begin{align}
            u^{linear}&=tanh(W^{u}u); v^{linear}=tanh(W^{v}v) \nonumber \\
			a_{ij} &= W_s(u_{i}^{linear} - {v_j^{linear}}^T) \\
            w_i &= softmax(a_i)  \\
            v^{weighted}&= w_i v \\
            u_{i}^{new}&=[u_i; v^{weighted}] \nonumber \label{conflict} \\
			\end{align}
			
where $W_s$ $\in$ $\mathbb{R}^{H \times 1}$ is a parameter that we introduce to provide a weight for the pair. The two word representations $u_i$ and $v_j$ are projected to a space where their element wise difference can be used to model their dissimilarity and softmax applied on them can produce high probability to more dissimilar word pairs.

It is good to note that conflict suffers from the same limitation that attention suffers from. This is when a pair of sentences are highly matching especially with multiple associations. But when the two methods work together, each compensates for the other's shortcomings.

\section{Combination of attention and conflict}

We used two weighted representations of $v$ using weights of attention and conflict as computed in Eqn. (4) and (8) respectively. Our final representation of a word representation $u_i$ conditioned on $v$ can be expressed as:
\begin{align}
    u_{i}^{new}&=[u_i;v_{weighted}^A;v_{weighted}^C] \\
\end{align}
where A and C denote that they are from attention and conflict models respectively.

\subsection{Relation to Multi-Head attention}
Multi-head attention, as introduced in \cite{DBLP:conf/nips/VaswaniSPUJGKP17}, computes multiple identical attention mechanism parallelly on multiple linear projections of same inputs. The parameters of each attention and projections are different in each head. Finally, they concatenate all the attentions which is similar to how we concatenate conflict and attention. However, they use dot-product to compute each of the attention.

Our combined model that contains both attention and conflict can be thought of as a 2-head attention model but both heads are different. Our conflict head explicitly captures difference between the inputs.

\section{Visualizing attention and conflict}
We observe how our conflict model learns the dissimilarities between word representations. We achieve that by visualizing the heatmap of the weight matrix $w$ for both attention and conflict from eqns. (3) and (8). While attention successfully learns the alignments, conflict matrix also shows that our approach models the contradicting associations like "animal" and "lake" or "australia" and "world". These two associations are the unique pairs which are instrumental in determining that the two queries are not similar.

\section{The model}
We create two models both of which constitutes of three main parts: encoder, interaction and classifier and take two sequences as input. Except interaction, all the other parts are exactly identical between the two models. The encoder is shared among the sequences simply uses two stacked GRU layers. The interaction part consists of \textit{only attention} for one model while for the another one it consists of \textit{attention and conflict combined} as shown in (eqn.11) . The classifier part is simply stacked fully-connected layers. Figure 3 shows a block diagram of how our model looks like.

\begin{figure}[h]
  \centering
  \includegraphics[width=\linewidth,height=3in]{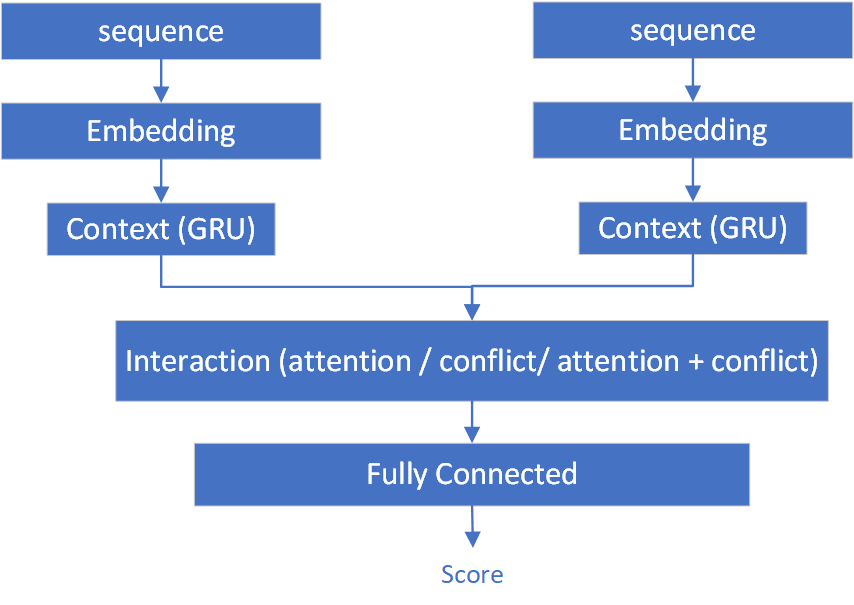}
  \caption{Generic Model containing interaction layer. We use attention, conflict or conjunction of attention and conflict as the interaction layer.}
\end{figure}

\section{Empirical Evaluation}
\subsection{Task 1: Quora Duplicate Question Pair Detection}
The dataset \footnote{https://data.quora.com/First-Quora-Dataset-Release-Question-Pairs} includes pairs of questions labelled as 1 or 0 depending on whether a pair is duplicate or not respectively. This is a popular pair-level classification task on which extensive work has already been done before like \cite{DBLP:conf/emnlp/TomarDTUD17}, \cite{DBLP:conf/ijcai/DuanCCWZZ18}. For this task, we make the output layer of our model to predict two probabilities for non-duplicate and duplicate. We sample the data from the original dataset so that it contains equal positive and negative classes. Original dataset has some class imbalance but for sake simplicity we don't consider it. The final data that we use has roughly 400,000 question pairs and we split this data into train and test using 8:2 ratio.

We train all our models for roughly 2 epochs with a batch size of 64. We use a hidden dimension of 150 throughout the model. The embedding layer uses ELMO \cite{DBLP:conf/naacl/PetersNIGCLZ18} which has proven to be very useful in various downstream language understanding tasks. Our FC layers consists of four dense layers with $tanh$ activation after each layer. The dropout rate is kept as 0.2 for every recurrent and FC linear layers. We use Adam optimizer in our experiment with \textit{epsilon}=1e-8, \textit{beta}=0.9 and learning rate=1e-3.

\begin{figure}[h]
  \centering
  \includegraphics[width=\linewidth,height=3in]{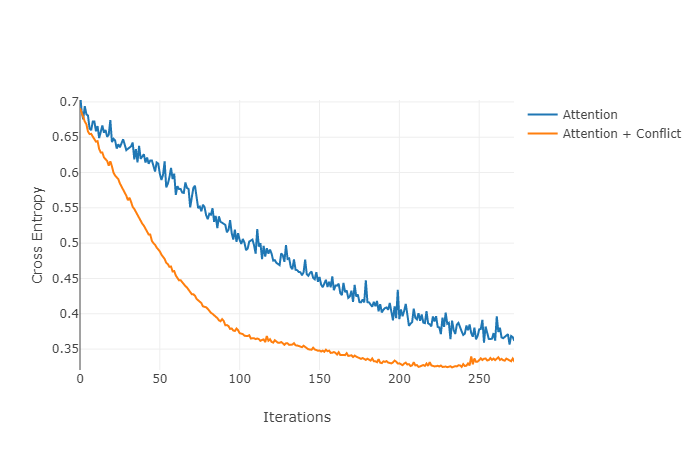}
  \caption{Training loss curve for Task 1}
\end{figure}

\begin{figure}[h]
  \centering
  \includegraphics[width=\linewidth,height=3in]{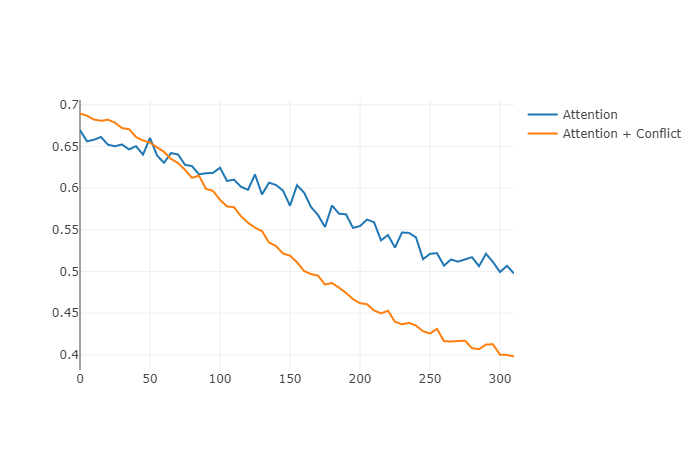}
  \caption{Training loss curve for Task 2}
\end{figure}

\subsection{Task 2: Ranking questions in Bing's People Also Ask}

People Also Ask is a feature in Bing search result page where related questions are recommended to the user. User may click on a question to view the answer. Clicking is a positive feedback that shows user's interest in the question. We use this click logs to build a question classifier using the same model in Figure 3. The problem statement is very similar to \cite{DBLP:conf/mm/ChenSLLH16} where they use logistic regression to predict whether an user would click on ad. Our goal is to classify if a question is potential high-click question or not for a given query. For this, we first create a labelled data set using the click logs where any question having CTR lower than 0.3 is labelled as 0 and a question having CTR more than 0.7 as 1.

Our final data resembles that of a pair-level classifier, as in Task 1, where user query and candidate questions are input. With these data set, we train a binary classifier to detect high-click and low-click questions.

\section{Results and Discussion}

\subsection{Quantitative Analysis}

For both tasks, we compute classification accuracy using three model variants and report the results in Table 1 and Table 2. We observe that model with both attention and conflict combined gives the best results.

We also show the training loss curve for both the models having attention and attention combined with conflict respectively. Figure 4 and 5 shows these curves for Task 1 and Task 2 respectively. The curves are smoothed using moving average having an window size of 8. We notice that the conflict model has much steeper slope and converges to a much better minima in both the tasks. It can also be noticed that in the training procedure for the model which has both attention and conflict, the updates are much smoother.

\begin{table}
  \caption{Result on Quora Dataset}
  \label{tab:freq}
  \begin{tabular}{ccl}
    \toprule
    Model&Accuracy&Cross-Entropy\\
    \midrule
    Model with only attention & 80.5& 0.34\\
    Model with only conflict & 80& 0.32\\
    Model with attention and conflict & \textbf{84.5} & \textbf{0.285}\\
  \bottomrule
\end{tabular}
\end{table}

\begin{table}
  \caption{Result on PAA click data}
  \label{tab:freq}
  \begin{tabular}{ccl}
    \toprule
    Model&Accuracy&Cross-Entropy\\
    \midrule
    Model with only attention & 85& 0.28\\
    Model with only conflict & 85.02& 0.276\\
    Model with attention and conflict & \textbf{88} & \textbf{0.25}\\
  \bottomrule
\end{tabular}
\end{table}

\subsection{Qualitative Comparison}
We also show qualitative results where we can observe that our model with attention and conflict combined does better on cases where pairs are non-duplicate and has very small difference. We have observed that the conflict model is very sensitive to even minor differences and compensates in such cases where attention poses high bias towards similarities already there in the sequences. \\

\textbf{Sequence 1}: What are the best ways to learn French ?

\textbf{Sequence 2}: How do I learn french genders ?

\textbf{Attention only}: 1

\textbf{Attention+Conflict}: 0

\textbf{Ground Truth}: 0 \\

\textbf{Sequence 1}: How do I prevent breast cancer ?

\textbf{Sequence 2}: Is breast cancer preventable ?

\textbf{Attention only}: 1

\textbf{Attention+Conflict}: 0

\textbf{Ground Truth}: 0

We provide two examples with predictions from the models with only attention and combination of attention and conflict. Each example is accompanied by the ground truth in our data.

\subsection{Analyzing the gains}
We analyzed the gains in Task 1 which we get from the attention-conflict model in order to ensure that they are not due to randomness in weight initialization or simply additional parameters. We particularly focused on the examples which were incorrectly marked in attention model but correctly in attention-conflict model. We saw that 70\% of those cases are the ones where the pair was incorrectly marked as duplicate in the previous model but our combined model correctly marked them as non-duplicate.

\section{Conclusion}
In this work, we highlighted the limits of attention especially in cases where two sequences have a contradicting relationship based on the task it performs. To alleviate this problem and further improve the performance, we propose a conflict mechanism that tries to capture how two sequences repel each other. This acts like the inverse of attention and, empirically, we show that how conflict and attention together can improve the performance.

Future research work should be based on alternative design of conflict mechanism using other difference operators other than element wise difference which we use.

\bibliographystyle{named}
\bibliography{ijcai19}

\end{document}